\documentclass{article}

\usepackage{graphicx}
\usepackage{booktabs}
\usepackage{multirow}
\usepackage{amsmath}
\usepackage{amssymb}
\usepackage{subcaption}
\usepackage[dvipsnames]{xcolor}
\usepackage{url}

\usepackage[preprint]{corl_2026} 

\title{AIR-VLA+: Decoupling Movement and Manipulation via Cascaded Dual-Action Decoders with Asymmetric MoE for Aerial Robots}

\author{
  \textbf{Jianli Sun$^{1}$} \quad
  \textbf{Bin Tian$^{1}$} \quad
  \textbf{Qiyao Zhang$^{2}$} \quad
  \textbf{Zijian Liu$^{3}$} \\
  \textbf{Yutong Wang$^{1}$} \quad
  \textbf{Zhiyong Cui$^{4}$} \quad
  \textbf{Bai Li$^{5}$} \quad
  \textbf{Yisheng Lv$^{1}$} \quad
  \textbf{Yonglin Tian$^{1}$}\thanks{Corresponding author. Email: \texttt{yonglin.tian@ia.ac.cn}} \quad
  \\
  \textmd{$^{1}$ The Institute of Automation, Chinese Academy of Sciences} \\
  \textmd{$^{2}$ School of Automation, Beijing Institute of Technology} \\
  \textmd{$^{3}$ College of Automotive and Energy Engineering, Tongji University} \\
  \textmd{$^{4}$ School of Transportation Science and Engineering, Beihang University} \\
  \textmd{$^{5}$ Information Science, East China Normal University}
}

\begin{document}
\maketitle


\begin{abstract}
Aerial manipulation systems have long suffered from representation coupling in end-to-end control, as platform-level Unmanned Aerial Vehicle (UAV) movement and end-effector-level arm manipulation differ substantially in action scale, dynamics, and control objectives. In this paper, we propose AIR-VLA+, a flow matching action generation architecture specifically designed for aerial manipulation, featuring cascaded dual-action decoders and an asymmetric feature-level Mixture of Experts (MoE). We construct cascaded manipulation and movement decoders, allowing the UAV to unidirectionally observe the manipulator's intent during movement to achieve workflow coordination, while isolating the impact of UAV movement information backpropagation on arm manipulation stability. Addressing the characteristic that UAV movement is highly dependent on high-level semantics and responsible for task state transitions in aerial manipulation, we design an input feature enhancement module for the UAV movement decoder. This module introduces an implicit visual grasp projector to perceive the interaction state between the gripper and the object, and injects compressed global semantic features. Within the UAV movement decoder, we deploy an implicit MoE architecture, enabling different movement experts to spontaneously exhibit capacity inclinations for various task stages during training. Through dense soft blending computation on the feature manifold, the UAV movement is endowed with stronger task-stage adaptability. Experiments on the standardized AIR-VLA benchmark demonstrate that our method comprehensively surpasses all baselines with an overall average score of 48.0. The overall task completion score improves by 80.2\% compared to the single-head $\pi_{0.5}$ policy, effectively mitigating the heterogeneous coordinated control conflicts of composite robots.
\end{abstract}



\section{Introduction}

Empowering physical entities with multimodal large models is a core pathway toward Artificial General Intelligence (AGI) and Embodied AI. Driven by the surging demand for complex long-horizon interactions, Vision-Language-Action (VLA) models have become mainstream. Particularly in continuous trajectory generation, flow matching has demonstrated superior convergence speed and trajectory smoothness compared to traditional diffusion models when modeling high-dimensional action spaces. The rise of this end-to-end architecture enables composite robots (e.g., aerial manipulation systems) to directly map natural language instructions into high-dimensional, closed-loop continuous action sequences.

However, directly applying existing VLA models to aerial manipulation systems often encounters highly heterogeneous action feature response requirements: the Unmanned Aerial Vehicle (UAV) focuses on macroscopic, large-scale movement (where the action output consists of relative positional changes), whereas the manipulator focuses on microscopic, local 3D high-precision alignment (where the action output consists of joint angles and gripper states). Traditional models force these two into joint optimization within a single shared latent space, triggering severe representation coupling and coordination conflicts. At the physical execution level, this directly manifests as UAV drifting and jittering, as well as uncoordinated transitions across task stages.

In aerial manipulation tasks, the UAV must execute large-scale movements after identifying the target object, hover steadily while the manipulator interacts with the object, and proceed to the next target only after confirming a successful grasp. Conversely, the manipulator strictly focuses on the precise grasping of the target object. Due to this divergence in task objectives, the UAV is highly dependent on continuous semantic understanding during the workflow to clarify target objects across different stages, deciding whether to hover or execute large-scale movements based on the manipulator's interaction state. Meanwhile, the manipulator predominantly emphasizes precise physical manipulation.

Addressing the characteristics of aerial manipulation tasks, we propose the \textbf{AIR-VLA+} architecture, realizing a system that is “cognitively highly coordinated and physically absolutely decoupled." We achieve action-level decoupling between the UAV and the manipulator through a cascaded dual-action decoder structure. We design an input feature enhancement module for the UAV movement decoder, introducing a miniature implicit nonlinear perception network to mine the physical contact representations between the gripper and the object, which are crucial to the task state. Furthermore, we incorporate a global semantic information compression module to maintain the UAV movement's understanding of task semantics, and employ a unidirectional information transmission cascade channel from the arm manipulation decoder to the UAV movement decoder. This enhances the workflow coordination capability between the UAV and the manipulator while protecting the pure inverse kinematics mapping of the manipulator. Finally, we adopt a Feature-Level Mixture of Experts (MoE) architecture in the UAV movement decoder. Through dense soft blending on the feature manifold, this enhances the UAV's flexibility during task stage transitions and pre-resolves the action feature conflicts between the UAV and the manipulator.

The core contributions of this paper are summarized as follows:
\begin{itemize}
    \item We propose \textbf{AIR-VLA+}, an asymmetric feature-level MoE flow matching action generation architecture featuring cascaded dual-action decoders tailored to the heterogeneous characteristics of aerial manipulation. This architecture greatly mitigates the performance degradation triggered by action representation coupling and significantly enhances the coordination level between the UAV and the manipulator.
    \item We design an input feature enhancement module for the UAV movement decoder to extract global semantic information and implicit visual grasp representations. This module resolves the issues of cross-modal feature submergence and shortcut learning, improves the UAV's movement adaptability during task transitions, and achieves efficient closed-loop coordination between the UAV and the manipulator.
    \item Comprehensive evaluations on the standardized AIR-VLA benchmark demonstrate that our method establishes state-of-the-art (SOTA) performance, achieving more stable flight, more precise manipulation, safer interactions, and superior task completion rates.
\end{itemize}

\begin{figure}[htbp]
    \centering
    \includegraphics[width=0.9\textwidth]{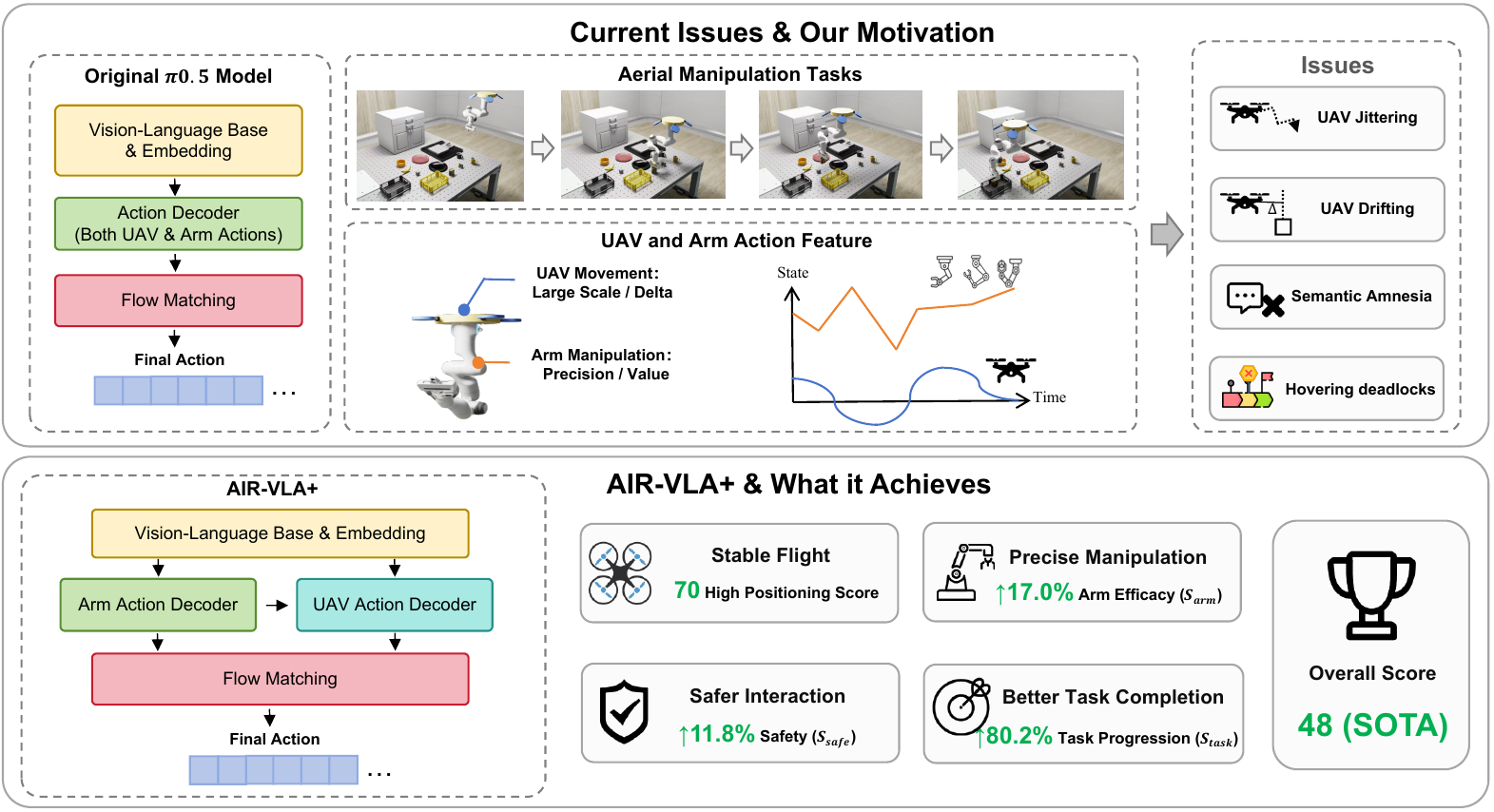}
    \caption{\textbf{Overview of the motivation and achievement of AIR-VLA+.}}
    \label{fig:method_overview}
\end{figure}


\section{Related Works}

UAVs have achieved large-scale applications in various fields. However, constrained by their mechanical design, traditional single flight platforms struggle to accomplish complex tasks such as pick and put operations. To overcome this limitation, aerial manipulator systems have emerged as a highly promising research hotspot.

In terms of mechanical structure, researchers have enhanced manipulation flexibility through bionic and cooperative designs. Examples include a lightweight robotic manipulator mimicking the hind limbs of an owl to reduce platform payload \cite{Liu2024}, a continuum manipulator offering excellent compliant manipulation for complex targets \cite{Peng2025}, and dual-arm cooperative structures that efficiently execute multi-step collaborative manipulation tasks \cite{Ghorbani2023}. 

Concurrently, advanced control methods have significantly improved manipulation accuracy and stability. These include refined anti-disturbance controller systems \cite{Liu2024}, adaptive prescribed performance control \cite{9812607}, hierarchical control strategies \cite{deshmukh2025globalendeffectorposecontrol}, and image-based visual impedance force control \cite{9750110}. Furthermore, techniques like coupled motion planning, dynamic compensation control, and whole-body integrated motion planning have further optimized overall system performance \cite{10547187}.

Despite these advances \cite{10943237, carvajal2024multitask, 11219352}, existing systems primarily focus on optimizing control metrics while lacking a global understanding of complex scenarios. An effective mechanism for efficiently translating natural language into continuous robotic action sequences is still missing. This insufficient generalization in dynamic environments highlights the urgent need to introduce multimodal intelligence technology.

VLA models realize a multimodal end-to-end unified architecture that enables embodied agents to execute physical interaction actions based on natural language. Compared with traditional modular models, this architecture possesses stronger semantic understanding and scene generalization abilities. Since 2022, models ranging from RT-1 \cite{brohan2023rt1roboticstransformerrealworld} and OpenVLA \cite{kim2024openvlaopensourcevisionlanguageactionmodel} to the $\pi$ series (\cite{black2026pi0visionlanguageactionflowmodel, black2025pi, intelligence2025pi06vlalearnsexperience, intelligence2026pi07steerablegeneralistrobotic}) have rapidly evolved, gradually making the leap from short-horizon tasks to strongly generalizable long-horizon tasks.

Relying on powerful multimodal fusion, VLA models have significantly promoted UAV intelligence. Applications span from coordinated navigation \cite{liu2025indooruavbenchmarkingvisionlanguageuav} and obstacle avoidance \cite{sun2026autoflyvisionlanguageactionmodeluav} to highly dynamic visual tracking \cite{zhang2026uavtrackvlaembodiedaerial} and direct kinematic mapping \cite{xu2026aerialvlavisionlanguageactionmodeluav}. In complex environments, VLAs have been optimized for highly dynamic racing \cite{serpiva2025racevlavlabasedracingdrone}, target interaction paradigms \cite{wang2025uavflowcolosseorealworldbenchmark}, and cognitive trajectory planning \cite{lykov2025cognitivedronevlamodelevaluation}. For resource-constrained onboard platforms, models utilizing high-fidelity datasets constructed via 3D Gaussian Splatting, combined with geometric safety correction, have achieved lightweight closed-loop navigation \cite{wu2025vlaanefficientonboardvisionlanguageaction}.

Although VLA technology has progressed in flight control, its research in aerial manipulation is still nascent and mostly limited to the direct transfer of foundation models. Sun et al. \cite{sun2026airvlavisionlanguageactionsystemsaerial} proposed AIR-VLA, verifying this transferability but revealing performance boundaries caused by the dynamic coupling of floating bases. Tucker et al. \cite{tucker2026pimakeflyphysicsguided} explored fixed-base VLA transfer to aerial platforms by injecting payload-aware physical constraints during the flow matching sampling stage. In summary, deep architectural improvements targeting the heterogeneous characteristics of aerial manipulator systems are still lacking. Therefore, to address the coupling issues between the base platform and manipulator joints, this paper proposes \textbf{AIR-VLA+}. It introduces asymmetric feature-level dynamic routing and a unidirectional information transmission cascade mechanism, achieving deep decoupling of the chassis and the manipulator without disrupting the continuity of the flow field, effectively filling the architectural gap in heterogeneous collaborative control.

\section{Methodology}

This section details the proposed Asymmetric Feature-Level MoE Flow Matching architecture. Aerial manipulation systems face severe challenges in long-horizon interaction tasks. To fundamentally resolve the representation coupling, we propose the AIR-VLA+ action generation architecture featuring cascaded dual-action decoders, realizing a system that is cognitively highly coordinated and physically absolutely decoupled (Figure \ref{fig:method_overview}).

\begin{figure}[htbp]
    \centering
    \includegraphics[width=0.9\textwidth]{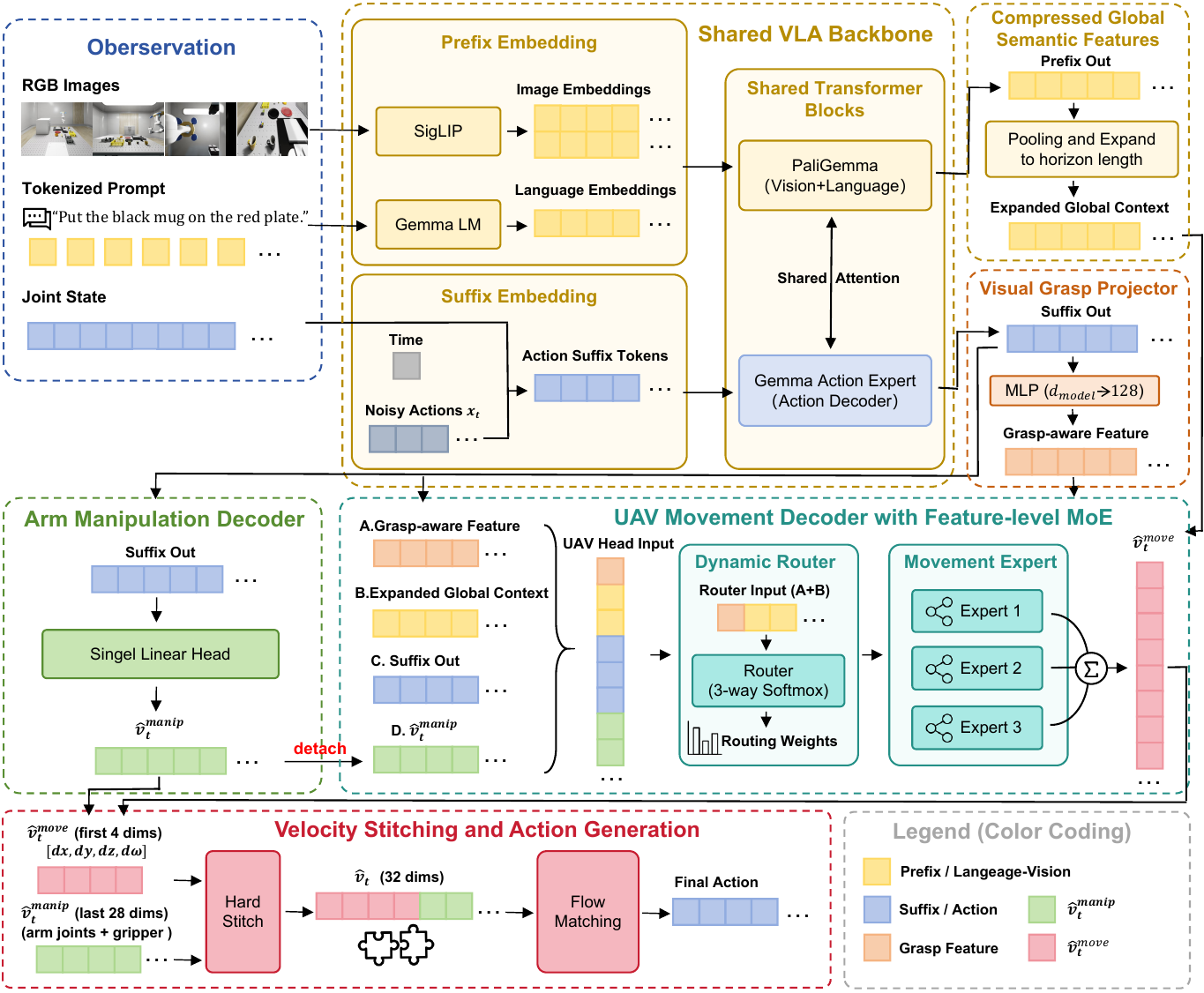}
    \caption{\textbf{Overview of the AIR-VLA+ asymmetric feature-level flow matching architecture.} The system comprises a shared multimodal pre-trained base, cascaded dual-action decoders, an input feature enhancement module, a unidirectional information transmission cascade channel, and an asymmetric feature-level MoE.}
    \label{fig:method_overview}
\end{figure}

\subsection{Preliminaries: Flow Matching for Continuous Action Generation}

Flow matching provides an efficient framework for training continuous-time generative models (e.g., Ordinary Differential Equations (ODEs)), demonstrating superior convergence and trajectory smoothness over Diffusion Models for high-dimensional robot action spaces.

We model the composite robot's action space as $x \in \mathbb{R}^{4 + d_{arm}}$. The first 4 dimensions represent the UAV's motion control (3D translational and 1D yaw velocities: $dx, dy, dz, dw$), while the remaining $d_{arm}$ dimensions denote the manipulator's joint angles and gripper state. Given a true action $x_0$ from the expert distribution $q(x_0)$ and noise $x_1 \sim \mathcal{N}(0, \mathbf{I})$, the conditional target vector field is a straight optimal transport trajectory:
\begin{equation}
    x_t = t \cdot \epsilon + (1-t) \cdot x_0, \quad t \in [0, 1]
\end{equation}
The objective is to fit the derivative field $v_t(x_t, t, c)$ by minimizing the Mean Squared Error (MSE):
\begin{equation}
    \mathcal{L}_{FM} = \mathbb{E}_{t, x_0, \epsilon} \left\| v_\theta(x_t, t, c) - (\epsilon - x_0) \right\|^2_2
\end{equation}
where $c$ is the conditional input. However, jointly optimizing the UAV and manipulator in a single shared vector field $v_\theta$ triggers task interference. The UAV focuses on macroscopic, large-scale movement (where the action output consists of relative positional changes), whereas the manipulator focuses on microscopic, local 3D high-precision alignment (where the action output consists of joint angles and gripper states). Topologically, the target manifold of the UAV movement is relatively smooth and continuous, whereas the manipulator's actions exhibit drastic and highly non-linear variations. If these topologically conflicting action features are jointly optimized within a single shared vector field, it triggers severe representation coupling and coordination conflicts, manifesting as UAV drifting and jittering during ODE integration.

\subsection{Cascaded Dual-Action Decoders}

To break through the bottleneck of joint optimization, we abandon the traditional symmetric design at the decoding end and construct cascaded manipulation and movement decoders. The overall architecture consists of a shared Vision-Language base (incorporating shared embeddings and shared Transformer blocks) and two specialized, asymmetric action decoders. 

The shared multimodal backbone first processes the inputs to extract foundational latent representations ($\mathbf{f}_{suffix}$). Subsequently, the arm manipulation decoder directly predicts the high-precision 3D alignment intent ($\hat{v}_{manip}$) from these shared features. To achieve workflow coordination, we employ a unidirectional information transmission cascade channel from the manipulation decoder to the UAV movement decoder. By deploying a stop-gradient operation (\texttt{.detach()}) on the manipulator's intent, this channel allows the UAV to unidirectionally observe the manipulator's intent. This ensures workflow coordination while strictly isolating the impact of UAV movement information backpropagation on arm manipulation stability, protecting the pure inverse kinematics mapping within the manipulator's manifold.

\subsection{UAV Movement Decoder with Input Feature Enhancement}

Addressing the characteristic that UAV movement is highly dependent on high-level semantics and responsible for task state transitions in aerial manipulation, we design an input feature enhancement module for the UAV movement decoder.

\textbf{Compressed Global Semantic Features:} We employ Mean Pooling as an information bottleneck to compress the sequence output of prefix tokens (images and text) into a global intent vector. This feature is temporally broadcasted across the action sequence horizon to form compressed global semantic features ($C_{global}$), maintaining the UAV's continuous understanding of task instructions.

\textbf{Implicit Visual Grasp Projector:} Aerial grasping requires definitive feedback on the contact state. We introduce a miniature implicit perception network (a 2-layer multi-layer perceptron (MLP)) to mine physical contact representations from the shared base visual features ($\mathbf{f}_{suffix}$). This squeezes the dimension to emerge with physical state discriminability, providing the visual grasp features ($\mathbf{f}_{grasp}$).

Ultimately, the enhanced input for the UAV movement decoder ($\mathbf{f}_{move\_input}$) is constructed by concatenating the shared foundational visual features, the compressed global semantic features, the visual grasp features, and the isolated manipulator intent:
\begin{equation}
    \mathbf{f}_{move\_input} = \text{Concat}(\left[ \mathbf{f}_{suffix},\; C_{global},\; \mathbf{f}_{grasp},\; \text{sg}(\hat{v}_{manip}) \right])
\end{equation}

\subsection{Asymmetric Feature-Level MoE}

Within the UAV movement decoder, we deploy an implicit Feature-Level MoE architecture to process $\mathbf{f}_{move\_input}$. The routing network (Router) acts as a decision bottleneck, computing activation weights for the experts based on the global semantics and grasp states:
\begin{equation}
    w_i = \text{Softmax}(\text{MLP}_{router}([C_{global},\; \mathbf{f}_{grasp}]))_i
\end{equation}
The parallel sub-expert networks ($E_1, \dots, E_K$) independently calculate implicit features, completing dense soft blending at the feature level:
\begin{equation}
    \hat{v}_{move} = \sum_{i=1}^K w_i \cdot E_i(\mathbf{f}_{move\_input})
\end{equation}
This computation on the feature manifold endows the UAV movement with stronger task-stage adaptability and mitigates chassis oscillation during task transitions.

\subsection{Action Space Hard Stitching and Flow Matching Integration}

After obtaining the derivative field $\hat{v}_{move}$ and the manipulation intent $\hat{v}_{manip}$, we execute physical hard stitching in the action space before the ODE solver. To preserve the pure kinematic mapping manifold, we eschew residual addition for strict dimension index $j$ slicing:
\begin{equation}
    \hat{v}_{t}^{(j)} = 
    \begin{cases} 
        \hat{v}_{move}^{(j)}, & j \in \{1, 2, 3, 4\} \\
        \hat{v}_{manip}^{(j)}, & j > 4 
    \end{cases}
\end{equation}
The first 4 dimensions ($dx, dy, dz, dw$) are taken over by the movement MoE; the remaining dimensions (manipulator joints and gripper) are governed by the arm manipulation decoder. During training, the unified $\mathcal{L}_{FM}$ loss backpropagates independently along this definitive physical boundary. 

During inference, this stitched global derivative field $\hat{v}_t$ guides the flow matching ODE solver. Starting from pure Gaussian noise $x_1 \sim \mathcal{N}(0, \mathbf{I})$, the solver integrates along the vector field over the continuous time horizon $t \in [1, 0]$ to iteratively denoise the state. This process ultimately yields the highly coordinated and physically decoupled continuous action sequence $x_0 \in \mathbb{R}^{4 + d_{arm}}$, which is directly deployed to execute closed-loop aerial manipulation tasks.

\section{Experiments}
\label{sec:experiments}

To systematically evaluate the effectiveness of the proposed architecture in Aerial Manipulation Systems, this section aims to answer the following core research questions (RQs):
\begin{itemize}
    \item \textbf{RQ1 (System Performance)}: Compared to action-level blending or hard stitching, can the asymmetric feature-level MoE effectively mitigate the representation coupling bottleneck and achieve superior performance in complex tasks?
    \item \textbf{RQ2 (Module Contribution)}: What are the independent contributions of components like the input feature enhancement module and the unidirectional information transmission cascade channel?
    \item \textbf{RQ3 (Interpretability)}: Do the implicit expert networks spontaneously emerge cognitive specialization and state estimation capabilities during end-to-end optimization? (Detailed experimental results and analyses are deferred to the Appendix.)
\end{itemize}

\subsection{Experimental Setup and Evaluation Metrics}

This study utilizes the \textbf{AIR-VLA} \cite{sun2026airvlavisionlanguageactionsystemsaerial} benchmark, which is specifically designed for aerial manipulation. All experiments are conducted within Isaac Sim, controlling a quadrotor UAV equipped with a 7-DoF manipulator (constituting a 12-DoF control space) and adhering to standard multi-view perception configurations. This benchmark encompasses four core task suites: Base Manipulation, Object \& Spatial, Semantic Understanding, and Long-Horizon. It provides a comprehensive evaluation of aerial manipulation performance across four dimensions: UAV positioning, arm manipulation efficacy, safety, and overall task completion.

\subsection{Main Results: Architecture Comparison (RQ1)}

We compare Feature-MoE with traditional architectures (ACT, Diffusion Policy) and flow matching large models ($\pi_0\text{-FAST}, \pi_0, \pi_{0.5}$). To verify internal architectural gains, we use $\pi_{0.5}$ as the base and compare the single head, Soft MoE, Hard MoE, and our method (Table~\ref{tab:comprehensive_results}).

\textbf{Quantitative Analysis:} Traditional policy models lack cross-entity pre-training and perform extremely poorly. Flow matching bases (like $\pi_0$), despite having generalization priors, exhibit obvious bottlenecks in heterogeneous action domains. The \textbf{Single Head ($\pi_{0.5}$)} baseline is constrained by representation coupling between the macroscopic UAV movement and the microscopic arm manipulation. Conventional schemes attempting to forcibly decouple at the action end all failed: \textbf{Action-Level Soft MoE} directly averages the derivative field $v_t$, destroying ODE trajectory continuity and leading to severe jitter and plummeting safety; \textbf{Action-Level Hard MoE} triggers solver and model collapse entirely due to step jumping.

In contrast, \textbf{Ours (Feature-MoE)}, via dense soft blending on the feature manifold and physical hard isolation at the action output end, comprehensively surpasses all baselines with an overall score of \textbf{48.0}. Compared to the single-head $\pi_{0.5}$ model, its task completion score improves by 80.2\%, achieving success rate leaps of 200\% and 150\% in base manipulation and long-horizon tasks, effectively mitigating the heterogeneous coordinated control conflicts.

\begin{table*}[t] 
\centering 
\caption{\textbf{Detailed performance evaluation of models across four task suites and overall average.} The table displays normalized scores for sub-metrics and weighted total scores for each model under different task types. Weights are set as $w_{pos}=0.25, w_{arm}=0.25, w_{safe}=0.10, w_{task}=0.40$. Abbreviations: \textbf{Pos}: Base Positioning Accuracy ($S_{pos}$); \textbf{Arm}: Manipulator Efficacy ($S_{arm}$); \textbf{Safe}: Environmental Safety ($S_{safe}$); \textbf{Task}: Task Progression ($S_{task}$); \textbf{Tot}: Weighted Total Score ($S_{total}$). Best results are highlighted in bold.} 
\label{tab:comprehensive_results} 
\renewcommand{\arraystretch}{1.1} 
\setlength{\tabcolsep}{1.5pt} 
\resizebox{\textwidth}{!}{%
\begin{tabular}{l ccccc ccccc ccccc ccccc ccccc} 
\toprule 
\multirow{3}{*}{\textbf{Model}} & \multicolumn{5}{c}{\textbf{Base Manipulation}} & \multicolumn{5}{c}{\textbf{Object \& Spatial}} & \multicolumn{5}{c}{\textbf{Semantic Understanding}} & \multicolumn{5}{c}{\textbf{Long-Horizon}} & \multicolumn{5}{c}{\textbf{Overall Average}} \\ 
\cmidrule(lr){2-6} \cmidrule(lr){7-11} \cmidrule(lr){12-16} \cmidrule(lr){17-21} \cmidrule(lr){22-26} 
 & Pos & Arm & Safe & Task & \textbf{Tot} & Pos & Arm & Safe & Task & \textbf{Tot} & Pos & Arm & Safe & Task & \textbf{Tot} & Pos & Arm & Safe & Task & \textbf{Tot} & Pos & Arm & Safe & Task & \textbf{Tot} \\ 
\midrule 
ACT  
 & 0.0 & 20.0 & \textbf{100.0} & 0.0 & 15.0  
 & 0.0 & 16.0 & \textbf{100.0} & 0.0 & 14.0  
 & 0.0 & 16.0 & \textbf{100.0} & 0.0 & 14.0  
 & 0.0 & 10.0 & \textbf{100.0} & 0.0 & 12.5  
 & 0.0 & 16.5 & \textbf{100.0} & 0.0 & 13.9 \\ 

Diffusion Policy 
 & 0.0 & 20.0 & \textbf{100.0} & 0.0 & 15.0  
 & 0.0 & 16.0 & \textbf{100.0} & 0.0 & 14.0  
 & 0.0 & 16.0 & \textbf{100.0} & 0.0 & 14.0  
 & 0.0 & 10.0 & \textbf{100.0} & 0.0 & 12.5  
 & 0.0 & 16.5 & \textbf{100.0} & 0.0 & 13.9 \\ 

$\pi_0\text{-FAST}$ 
 & 27.5 & 23.9 & 98.1 & 0.0 & 22.7  
 & 5.1 & 18.9 & 98.3 & 0.0 & 15.8  
 & 6.5 & 16.9 & 99.3 & 0.0 & 15.8  
 & 41.0 & 32.6 & 99.4 & 0.0 & 28.3  
 & 15.0 & 20.8 & 98.5 & 0.0 & 18.8 \\ 

$\pi_{0}$ 
 & 56.6 & 48.0 & 87.5 & 8.0 & 38.1  
 & 58.1 & 38.2 & 75.4 & 2.0 & 32.4  
 & 55.7 & 36.1 & 85.1 & 2.00 & 32.3  
 & \textbf{65.1} & 38.5 & 85.9 & 2.0 & 32.3  
 & 58.8 & 42.0 & 83.5 & 3.50 & 34.5 \\ 
\midrule
Single Head ($\pi_{0.5}$) 
 & 78.8 & 49.9 & 87.2 & 11.0 & 45.3  
 & \textbf{70.9} & 44.1 & 81.2 & 13.5 & 42.3  
 & \textbf{65.4} & 44.7 & 82.7 & \textbf{11.0} & 40.2  
 & 62.4 & 40.8 & 84.8 & 7.0 & 37.1  
 & \textbf{71.3} & 45.9 & 84.7 & 10.6 & 42.0 \\  

Action-Level Soft MoE
 & 54.9 & 42.3 & 41.4 & 0.5 & 28.6
 & 64.1 & 38.9 & 47.2 & 6.5 & 33.1
 & 62.5 & 39.5 & 46.6 & 5.5 & 32.3
 & 52.1 & 42.4 & 29.5 & 9.5 & 30.4
 & 52.4 & 40.8 & 41.2 & 5.5 & 31.1 \\

Action-Level Hard MoE
 & 0.0 & 20.0 & \textbf{100.0} & 0.0 & 15.0
 & 0.4 & 19.8 & \textbf{100.0} & 0.0 & 15.1
 & 0.0 & 19.2 & \textbf{100.0} & 0.0 & 14.8
 & 1.2 & 17.4 & 99.5 & 0.3 & 14.7
 & 0.4 & 19.1 & 99.9 & 0.1 & 14.9 \\

\textbf{Ours (Feature-MoE)}
 & \textbf{87.0} & \textbf{68.8} & 99.2 & \textbf{33.0} & \textbf{62.1}
 & 68.3 & \textbf{49.4} & 87.2 & \textbf{15.6} & \textbf{44.4}
 & 65.0 & \textbf{45.1} & 96.4 & 10.0 & \textbf{41.2}
 & 59.1 & \textbf{60.0} & 96.1 & \textbf{17.5} & \textbf{44.1}
 & 70.0 & \textbf{53.7} & 94.7 & \textbf{19.1} & \textbf{48.0} \\
\bottomrule 
\end{tabular}%
} 
\end{table*}

\subsection{Ablation Study (RQ2)}

To verify the independent contributions of each component, we designed a bottom-up ablation study (Table~\ref{tab:ablation_results}).

\begin{table*}[t] 
\centering 
\caption{\textbf{Detailed Ablation Study.} Experiments verify the independent contributions of the compressed global semantic features, visual perception, the stop-gradient operation, and multi-expert routing modules.} 
\label{tab:ablation_results} 
\renewcommand{\arraystretch}{1.1} 
\setlength{\tabcolsep}{1.5pt} 
\resizebox{\textwidth}{!}{%
\begin{tabular}{l ccccc ccccc ccccc ccccc ccccc} 
\toprule 
\multirow{3}{*}{\textbf{Model Variants}} & \multicolumn{5}{c}{\textbf{Base Manipulation}} & \multicolumn{5}{c}{\textbf{Object \& Spatial}} & \multicolumn{5}{c}{\textbf{Semantic Understanding}} & \multicolumn{5}{c}{\textbf{Long-Horizon}} & \multicolumn{5}{c}{\textbf{Overall Average}} \\ 
\cmidrule(lr){2-6} \cmidrule(lr){7-11} \cmidrule(lr){12-16} \cmidrule(lr){17-21} \cmidrule(lr){22-26} 
 & Pos & Arm & Safe & Task & \textbf{Tot} & Pos & Arm & Safe & Task & \textbf{Tot} & Pos & Arm & Safe & Task & \textbf{Tot} & Pos & Arm & Safe & Task & \textbf{Tot} & Pos & Arm & Safe & Task & \textbf{Tot} \\ 
\midrule 
(a) Original $\pi_{0.5}$ Base
 & 78.8 & 49.9 & 87.2 & 11.0 & 45.3  
 & \textbf{70.9} & 44.1 & 81.2 & 13.5 & 42.3  
 & \textbf{65.4} & 44.7 & 82.7 & 11.0 & 40.2  
 & 62.4 & 40.8 & 84.8 & 7.0 & 37.1  
 & \textbf{71.3} & 45.9 & 84.7 & 10.6 & 42.0 \\  
(b) Multi-Head (Move+Manip)
 & 85.0 & 61.6 & 99.6 & 17.0 & 53.4
 & 27.9 & 25.8 & 93.7 & 4.0 & 24.4
 & 25.0 & 22.9 & 89.9 & 5.0 & 23.0
 & \textbf{63.7} & 51.6 & 92.7 & 20.0 & 46.1
 & 50.4 & 40.5 & 94.0 & 11.5 & 36.7 \\
(c) (b) + Global Semantics
 & 35.3 & 0.0 & \textbf{99.9} & 0.0 & 10.1
 & 13.4 & 13.0 & \textbf{96.6} & 0.0 & 16.3
 & 12.9 & 13.4 & 95.7 & 0.0 & 16.1
 & 53.9 & 41.8 & 94.4 & 5.0 & 35.4
 & 20.2 & 17.1 & \textbf{96.7} & 1.25 & 19.5 \\
(d) (b) + Vis Grasp Feat
 & 79.1 & 55.6 & 84.1 & 12.0 & 47.9
 & 30.5 & 29.1 & 91.6 & 2.0 & 24.9
 & 32.3 & 32.0 & 86.1 & 7.0 & 27.9
 & 71.4 & 51.5 & 95.9 & \textbf{20.5} & \textbf{48.5}
 & 53.3 & 42.1 & 91.9 & 10.3 & 37.2 \\
(e) (b) + Semantics + Grasp (w/o Detach) 
 & 74.8 & 59.5 & 97.9 & 13.0 & 48.6
 & 55.5 & 43.2 & 92.5 & 12.0 & 38.7
 & 55.8 & 42.4 & 93.8 & 9.0 & 37.5
 & 49.8 & 46.1 & 93.7 & 9.0 & 36.9
 & 59.0 & 47.8 & 94.5 & 10.8 & 40.4 \\
(f) (b) + Semantics + Grasp (w/ Detach) 
 & 85.9 & 62.3 & 98.1 & 18.0 & 54.1
 & 67.6 & \textbf{50.5} & 89.1 & \textbf{20.0} & \textbf{46.4}
 & 63.0 & 44.9 & 86.3 & 14.0 & \textbf{41.2}
 & 55.2 & 44.0 & \textbf{98.0} & 10.0 & 38.6
 & 67.9 & 50.4 & 92.9 & 15.5 & 45.1 \\
\midrule
\textbf{(g) Full Feature-MoE}
 & \textbf{87.0} & \textbf{68.8} & 99.2 & \textbf{33.0} & \textbf{62.1}
 & 68.3 & \textbf{49.4} & 87.2 & \textbf{15.6} & \textbf{44.4}
 & 65.0 & \textbf{45.1} & \textbf{96.4} & 10.0 & \textbf{41.2}
 & 59.1 & \textbf{60.0} & 96.1 & \textbf{17.5} & \textbf{44.1}
 & 70.0 & \textbf{53.7} & 94.7 & \textbf{19.1} & \textbf{48.0} \\
\bottomrule 
\end{tabular}%
} 
\end{table*}

\textbf{Module Contribution Analysis:} Compared to base (a), the pure dual-action decoder split (b) alleviates dynamic conflicts but severs cross-modal coordination, degrading spatial and semantic reasoning tasks. Injecting only global semantics and manipulator intent into the UAV decoder (c) causes Destructive Interference: lacking microscopic physical anchoring, macroscopic instructions overwhelm visual signals, plunging the model into Shortcut Learning. Conversely, introducing only visual grasp features (d) rebuilds temporal dependency but lacks macroscopic guidance, inducing Semantic Amnesia and deteriorating into a cognitively blind state machine. Furthermore, omitting the stop-gradient (\texttt{.detach()}) operation (e) allows UAV movement backpropagation to severely perturb the manipulator's latent manifold. Physically, this dynamic interference manifests as severe manipulator jitter, drastically reducing manipulation efficacy and success rates. Enforcing \texttt{.detach()} (f) successfully isolates this conflict and balances feature integration. Finally, the complete \textbf{Feature-MoE (g)} breaks static rigidity via asymmetric routing, utilizing dense soft blending to optimize task transition smoothness and achieve the best overall results.

\textbf{Sensitivity to the Number of Experts:} The ablation study on the number of UAV movement experts ($N$) (Table~\ref{tab:expert_number}) demonstrates that at $N=2$, residual representation coupling persists, failing to smoothly partition transition states and resulting in a degradation of core metrics. Conversely, at $N=4$, the expanded parameter space does not yield the anticipated gains; certain experts maintain persistently low activation weights and lack distinct features that adapt independently to task phases. This corroborates that aerial manipulation tasks inherently converge into three latent spaces, establishing $N=3$ as the optimal number of experts.

\begin{table}[htbp]
\centering
\caption{\textbf{Sensitivity analysis on the number of latent experts ($N$).} The table reports the performance of Feature-MoE variants with different expert capacities. Values in parentheses indicate the performance gap ($\downarrow \Delta$) compared to the optimal $N=3$ setting.}
\label{tab:expert_number}
\small 
\begin{tabular}{lccccc}
\toprule
\textbf{Capacity} & \boldmath{$S_{pos}$} & \boldmath{$S_{arm}$} & \boldmath{$S_{safe}$} & \boldmath{$S_{task}$} & \boldmath{$S_{total}$} \\
\midrule
$N=3$ (Ours) 
 & \textbf{70.0} & \textbf{53.7} & \textbf{94.7} & \textbf{19.1} & \textbf{48.0} \\ 
\midrule
$N=2$ Experts
 & 64.8 \scriptsize{\color{gray}($\downarrow$5.2)} 
 & 45.8 \scriptsize{\color{gray}($\downarrow$7.9)} 
 & 94.6 \scriptsize{\color{gray}($\downarrow$0.1)} 
 & 7.0 \scriptsize{\color{gray}($\downarrow$12.1)} 
 & 39.9 \scriptsize{\color{gray}($\downarrow$8.1)} \\
$N=4$ Experts
 & 67.1 \scriptsize{\color{gray}($\downarrow$2.9)} 
 & 48.5 \scriptsize{\color{gray}($\downarrow$5.2)} 
 & 94.7 
 & 11.5 \scriptsize{\color{gray}($\downarrow$7.6)} 
 & 43.0 \scriptsize{\color{gray}($\downarrow$5.0)} \\
\bottomrule
\end{tabular}
\end{table}

\subsection{Interpretability: Emergence of Experts (RQ3)}
This section investigates whether implicit expert networks spontaneously emerge physical cognition and functional division of labor under flow matching end-to-end optimization.

\begin{figure}[htbp]
    \centering
    \includegraphics[width=\textwidth]{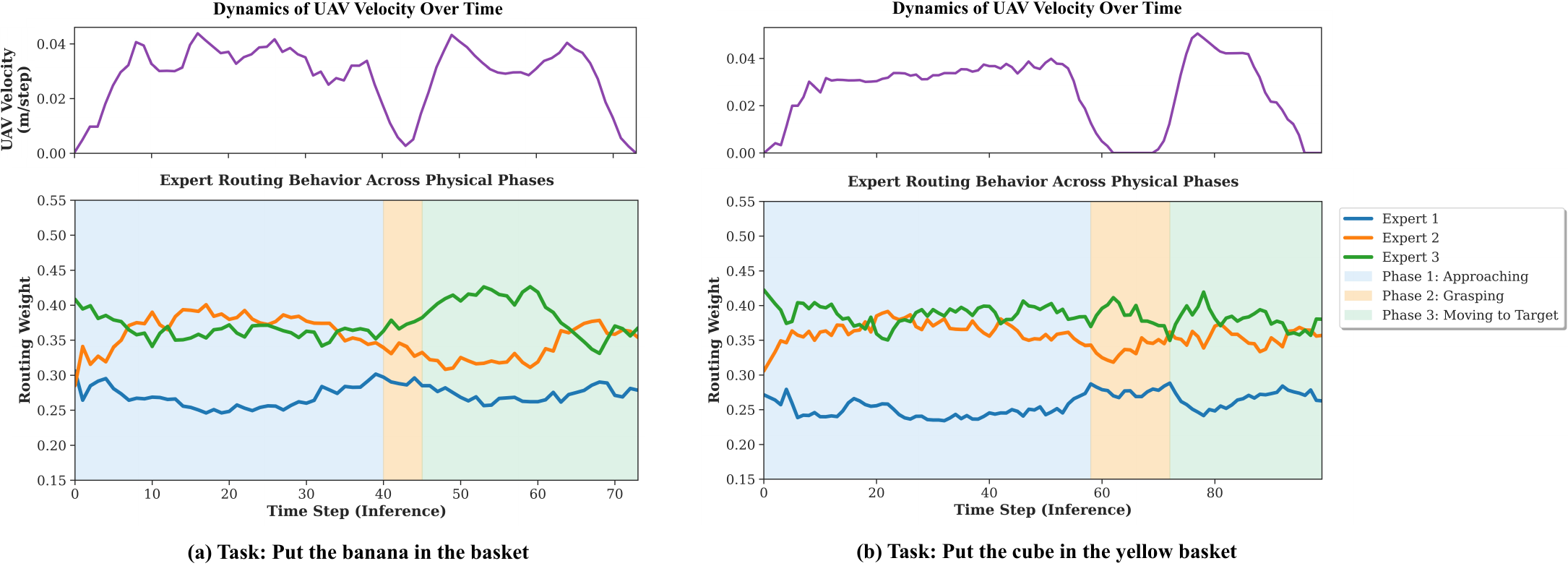}
    \caption{\textbf{Routing Dynamics in Long-Horizon Task.} Within a single task cycle, the three implicit experts demonstrate a highly logical temporal handover mechanism.}
    \label{fig:routing_dynamics}
\end{figure}

\textbf{Routing Dynamics:} As illustrated in Fig.~\ref{fig:routing_dynamics}, we track the evolution of expert weights throughout a single task cycle. Starting from an initial uniform distribution (0.333), the weights exhibit a significant phase transition after training: during the \textit{Approaching} stage, Expert 2's weight demonstrates an upward trend alongside UAV activation; during the \textit{Grasping} stage, Expert 1 reaches a local peak; whereas in the \textit{Moving to target} stage, Expert 3 leaps significantly and dominates. This trend intuitively reflects the Router's success in executing dynamic control based on task state transitions. In summary, Expert 1 exhibits a capacity inclination for the \textit{grasping} stage, Expert 2 for the \textit{approaching} stage, and Expert 3 for the \textit{moving to target} stage.

\textbf{Expert Isolation and Hard-Routing Analysis:} 
To verify the essence of physical decoupling, we conducted forced routing experiments (locking one-hot weights, Table~\ref{tab:expert_isolation}). We observed varying degrees of performance degradation when only a single expert was activated. Activating only Expert 1 results in the minimal overall performance drop, yet the UAV frequently exhibits sluggishness when moving towards the target object. Activating only Expert 2 or Expert 3 leads to significant performance degradation, specifically manifesting as the UAV maintaining excessively high velocities, frequently overshooting the target object, and failing to hover at the appropriate position. This phenomenon quantitatively corroborates that a single network struggles to balance the diverse dynamic requirements across all stages of an aerial manipulation task.

\begin{table}[htbp]
\centering
\caption{\textbf{Performance degradation under expert isolation (Hard-Routing).} The table shows the absolute scores when routing weights are forcibly locked to a single expert. Values in parentheses indicate the performance drop ($\downarrow \Delta$) or gain ($\uparrow \Delta$) compared to the dynamic Feature-MoE baseline.}
\label{tab:expert_isolation}
\small 
\begin{tabular}{lccccc}
\toprule
\textbf{Routing Condition} & \boldmath{$S_{pos}$} & \boldmath{$S_{arm}$} & \boldmath{$S_{safe}$} & \boldmath{$S_{task}$} & \boldmath{$S_{total}$} \\
\midrule
Dynamic (Ours) 
 & \textbf{70.0} & \textbf{53.7} & 94.7 & \textbf{19.1} & \textbf{48.0} \\ 
\midrule
Expert 1 Only \scriptsize{([1,0,0])}
 & 69.7 \scriptsize{\color{gray}($\downarrow$0.3)} 
 & 53.2 \scriptsize{\color{gray}($\downarrow$0.5)} 
 & 93.6 \scriptsize{\color{gray}($\downarrow$1.1)} 
 & 18.8 \scriptsize{\color{gray}($\downarrow$0.3)} 
 & 47.6 \scriptsize{\color{gray}($\downarrow$0.4)} \\
Expert 2 Only \scriptsize{([0,1,0])}
 & 67.2 \scriptsize{\color{gray}($\downarrow$2.8)} 
 & 48.6 \scriptsize{\color{gray}($\downarrow$5.1)} 
 & 94.9 \scriptsize{\color{gray}($\uparrow$0.2)} 
 & 10.8 \scriptsize{\color{gray}($\downarrow$8.3)} 
 & 42.7 \scriptsize{\color{gray}($\downarrow$5.3)} \\
Expert 3 Only \scriptsize{([0,0,1])}
 & 67.8 \scriptsize{\color{gray}($\downarrow$2.2)} 
 & 50.1 \scriptsize{\color{gray}($\downarrow$3.6)} 
 & \textbf{95.8} \scriptsize{\color{gray}($\uparrow$1.2)} 
 & 14.8 \scriptsize{\color{gray}($\downarrow$4.3)} 
 & 45.0 \scriptsize{\color{gray}($\downarrow$3.0)} \\
\bottomrule
\end{tabular}
\end{table}

\begin{figure*}[htbp]
    \centering
    \includegraphics[width=\textwidth]{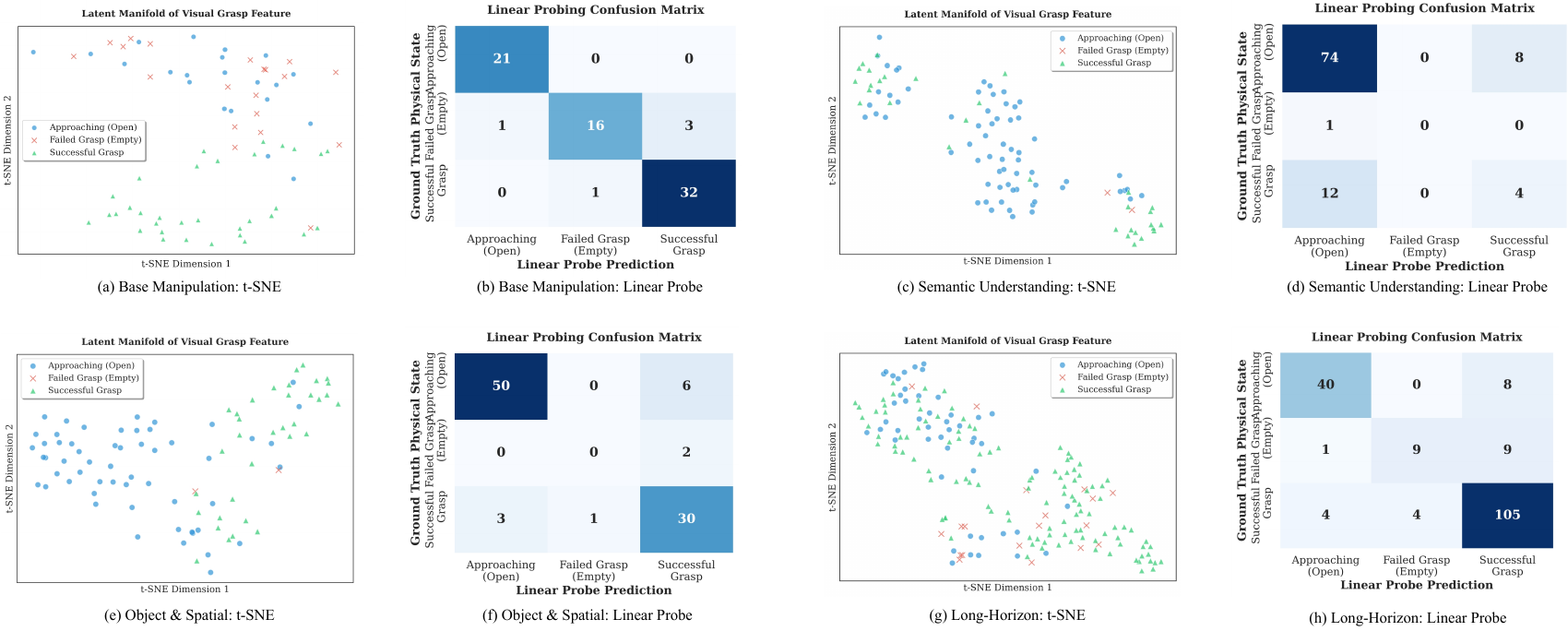}
    \caption{\textbf{Cross-Task Qualitative and Quantitative Analysis of Latent Manifold.} 
    The figure displays the t-SNE dimensionality reduction projections of 128-dimensional features and the cross-validation confusion matrices of pure Linear Probing across four tasks. Even facing severe reduction artifacts and visual overlap, the linear probe maintains extremely high physical state separation accuracy.}
    \label{fig:tsne_latent}
\end{figure*}

\textbf{Latent Manifold Analysis:}
To explore the latent manifold nature of the perceptor, we extracted hidden representations in each task suite and conducted t-SNE projections and Linear Probing tests (Fig.~\ref{fig:tsne_latent}). In simple \textbf{Base Manipulation}, features are highly linearly separable (93.24\% accuracy). Adversarial samples representing "gripper closed but not touching object" are accurately expelled from the "successful grasp" cluster by high-dimensional boundaries, confirming the model evades shortcut traps of merely fitting motion coordinates. Progressing to \textbf{Object}, \textbf{Long-Horizon}, and \textbf{Semantic} tasks, while the native manifold geometry becomes highly complex due to environmental noise (2D projections visually overlap), its linear classification accuracy in the native space remains high at 78\%-87\%. This ability to maintain a robust hyperplane under extremely dense semantic noise indicates the network does not mechanically memorize visual appearances but successfully extracts interaction semantics into the decision input of the UAV movement decoder.


\section{Conclusion}

To address representation coupling and coordinated control conflicts in aerial manipulation, we propose AIR-VLA+, a flow-matching architecture utilizing cascaded dual-action decoders and an asymmetric feature-level MoE. An input feature enhancement module introduces an implicit visual grasp projector for gripper-object perception and injects global semantics to maintain UAV task awareness. A stop-gradient unidirectional information transmission cascade channel enables the UAV to observe manipulator intent for coordination, while isolating movement backpropagation to preserve arm stability. Additionally, the UAV decoder's MoE employs dense soft blending on the feature manifold, enhancing task-stage adaptability without disrupting manipulator inverse kinematics. AIR-VLA benchmark experiments validate that our approach significantly improves coordination and achieves state-of-the-art performance, offering a novel multi-expert routing paradigm for composite robots.


\bibliography{example}  

@misc{lykov2025cognitivedronevlamodelevaluation,
      title={CognitiveDrone: A VLA Model and Evaluation Benchmark for Real-Time Cognitive Task Solving and Reasoning in UAVs}, 
      author={Artem Lykov and Valerii Serpiva and Muhammad Haris Khan and Oleg Sautenkov and Artyom Myshlyaev and Grik Tadevosyan and Yasheerah Yaqoot and Dzmitry Tsetserukou},
      year={2025},
      eprint={2503.01378},
      archivePrefix={arXiv},
      primaryClass={cs.RO},
      url={https://arxiv.org/abs/2503.01378}, 
}

@misc{liu2025indooruavbenchmarkingvisionlanguageuav,
      title={IndoorUAV: Benchmarking Vision-Language UAV Navigation in Continuous Indoor Environments}, 
      author={Xu Liu and Yu Liu and Hanshuo Qiu and Yang Qirong and Zhouhui Lian},
      year={2025},
      eprint={2512.19024},
      archivePrefix={arXiv},
      primaryClass={cs.RO},
      url={https://arxiv.org/abs/2512.19024}, 
}

@misc{serpiva2025racevlavlabasedracingdrone,
      title={RaceVLA: VLA-based Racing Drone Navigation with Human-like Behaviour}, 
      author={Valerii Serpiva and Artem Lykov and Artyom Myshlyaev and Muhammad Haris Khan and Ali Alridha Abdulkarim and Oleg Sautenkov and Dzmitry Tsetserukou},
      year={2025},
      eprint={2503.02572},
      archivePrefix={arXiv},
      primaryClass={cs.RO},
      url={https://arxiv.org/abs/2503.02572}, 
}

@misc{wang2025uavflowcolosseorealworldbenchmark,
      title={UAV-Flow Colosseo: A Real-World Benchmark for Flying-on-a-Word UAV Imitation Learning}, 
      author={Xiangyu Wang and Donglin Yang and Yue Liao and Wenhao Zheng and wenjun wu and Bin Dai and Hongsheng Li and Si Liu},
      year={2025},
      eprint={2505.15725},
      archivePrefix={arXiv},
      primaryClass={cs.RO},
      url={https://arxiv.org/abs/2505.15725}, 
}

@misc{wu2025vlaanefficientonboardvisionlanguageaction,
      title={VLA-AN: An Efficient and Onboard Vision-Language-Action Framework for Aerial Navigation in Complex Environments}, 
      author={Yuze Wu and Mo Zhu and Xingxing Li and Yuheng Du and Yuxin Fan and Wenjun Li and Zhichao Han and Xin Zhou and Fei Gao},
      year={2025},
      eprint={2512.15258},
      archivePrefix={arXiv},
      primaryClass={cs.RO},
      url={https://arxiv.org/abs/2512.15258}, 
}

@Article{Liu2024,
author={Liu, Qianyuan
and Liu, Yuhang
and Chen, Zeshuai
and Guo, Kexin
and Yu, Xiang
and Zhang, Youmin
and Guo, Lei},
title={A Compact Aerial Manipulator: Design and Control for Dexterous Operations},
journal={Journal of Intelligent {\&} Robotic Systems},
year={2024},
month={Apr},
day={26},
volume={110},
number={2},
pages={66},
issn={1573-0409},
doi={10.1007/s10846-024-02090-7},
url={https://doi.org/10.1007/s10846-024-02090-7}
}

@Article{Peng2025,
author={Peng, Rui
and Wang, Yu
and Lu, Minghao
and Lu, Peng},
title={A dexterous and compliant aerial continuum manipulator for cluttered and constrained environments},
journal={Nature Communications},
year={2025},
month={Jan},
day={21},
volume={16},
number={1},
pages={889},
issn={2041-1723},
doi={10.1038/s41467-024-55157-2},
url={https://doi.org/10.1038/s41467-024-55157-2}
}

@ARTICLE{9812607,
  author={Liang, Jiacheng and Chen, Yanjie and Wu, Yangning and Miao, Zhiqiang and Zhang, Hui and Wang, Yaonan},
  journal={IEEE Transactions on Automation Science and Engineering}, 
  title={Adaptive Prescribed Performance Control of Unmanned Aerial Manipulator With Disturbances}, 
  year={2023},
  volume={20},
  number={3},
  pages={1804-1814},
  doi={10.1109/TASE.2022.3186315}}

@ARTICLE{10943237,
  author={Zhang, Zhaopeng and Yu, Hai and Chai, Yi and Yang, Zhichao and Liang, Xiao and Fang, Yongchun and Han, Jianda},
  journal={IEEE/ASME Transactions on Mechatronics}, 
  title={An End-Effector-Oriented Coupled Motion Planning Method for Aerial Manipulators in Constrained Environments}, 
  year={2025},
  volume={30},
  number={6},
  pages={6027-6037},
  doi={10.1109/TMECH.2025.3550562}}

@Article{Ghorbani2023,
author={Ghorbani, Shahab
and Samadikhoshkho, Zahra
and Janabi--Sharifi, Farrokh},
title={Dual-arm aerial continuum manipulation systems: modeling, pre-grasp planning, and control},
journal={Nonlinear Dynamics},
year={2023},
month={Apr},
day={01},
volume={111},
number={8},
pages={7339-7355},
issn={1573-269X},
doi={10.1007/s11071-022-08212-w},
url={https://doi.org/10.1007/s11071-022-08212-w}
}

@misc{deshmukh2025globalendeffectorposecontrol,
      title={Global End-Effector Pose Control of an Underactuated Aerial Manipulator via Reinforcement Learning}, 
      author={Shlok Deshmukh and Javier Alonso-Mora and Sihao Sun},
      year={2025},
      eprint={2512.21085},
      archivePrefix={arXiv},
      primaryClass={cs.RO},
      url={https://arxiv.org/abs/2512.21085}, 
}

@ARTICLE{9750110,
  author={Xu, Mengxin and Hu, An and Wang, Hesheng},
  journal={IEEE Transactions on Automation Science and Engineering}, 
  title={Image-Based Visual Impedance Force Control for Contact Aerial Manipulation}, 
  year={2023},
  volume={20},
  number={1},
  pages={518-527},
  doi={10.1109/TASE.2022.3162207}}

@article{carvajal2024multitask,
title = {Multitask control of aerial manipulator robots with dynamic compensation based on numerical methods},
journal = {Robotics and Autonomous Systems},
volume = {173},
pages = {104614},
year = {2024},
issn = {0921-8890},
doi = {https://doi.org/10.1016/j.robot.2023.104614},
url = {https://www.sciencedirect.com/science/article/pii/S0921889023002531},
author = {Christian P. Carvajal and Gabriela M. Andaluz and Víctor H. Andaluz and Flavio Roberti and Guillermo Palacios-Navarro and Ricardo Carelli}
}

@ARTICLE{10547187,
  author={Wang, Meng and Lyu, Shangke and Liu, Qianyuan and Yang, Ziqi and Guo, Kexin and Yu, Xiang},
  journal={IEEE Transactions on Automation Science and Engineering}, 
  title={Precise End-Effector Control for an Aerial Manipulator Under Composite Disturbances: Theory and Experiments}, 
  year={2025},
  volume={22},
  number={},
  pages={4006-4021},
  doi={10.1109/TASE.2024.3406754}}

@ARTICLE{11219352,
  author={Deng, Weiliang and Chen, Hongming and Ye, Biyu and Chen, Haoran and Li, Ziliang and Lyu, Ximin},
  journal={IEEE Transactions on Robotics}, 
  title={Whole-Body Integrated Motion Planning for Aerial Manipulators}, 
  year={2025},
  volume={41},
  number={},
  pages={6661-6679},
  doi={10.1109/TRO.2025.3626619}}

@inproceedings{
black2025pi,
title={\${\textbackslash}pi\_\{0.5\}\$: a Vision-Language-Action Model with Open-World Generalization},
author={Kevin Black and Noah Brown and James Darpinian and Karan Dhabalia and Danny Driess and Adnan Esmail and Michael Robert Equi and Chelsea Finn and Niccolo Fusai and Manuel Y. Galliker and Dibya Ghosh and Lachy Groom and Karol Hausman and brian ichter and Szymon Jakubczak and Tim Jones and Liyiming Ke and Devin LeBlanc and Sergey Levine and Adrian Li-Bell and Mohith Mothukuri and Suraj Nair and Karl Pertsch and Allen Z. Ren and Lucy Xiaoyang Shi and Laura Smith and Jost Tobias Springenberg and Kyle Stachowicz and James Tanner and Quan Vuong and Homer Walke and Anna Walling and Haohuan Wang and Lili Yu and Ury Zhilinsky},
booktitle={9th Annual Conference on Robot Learning},
year={2025},
url={https://openreview.net/forum?id=vlhoswksBO}
}

@misc{black2026pi0visionlanguageactionflowmodel,
      title={$\pi_0$: A Vision-Language-Action Flow Model for General Robot Control}, 
      author={Kevin Black and Noah Brown and Danny Driess and Adnan Esmail and Michael Equi and Chelsea Finn and Niccolo Fusai and Lachy Groom and Karol Hausman and Brian Ichter and Szymon Jakubczak and Tim Jones and Liyiming Ke and Sergey Levine and Adrian Li-Bell and Mohith Mothukuri and Suraj Nair and Karl Pertsch and Lucy Xiaoyang Shi and James Tanner and Quan Vuong and Anna Walling and Haohuan Wang and Ury Zhilinsky},
      year={2026},
      eprint={2410.24164},
      archivePrefix={arXiv},
      primaryClass={cs.LG},
      url={https://arxiv.org/abs/2410.24164}, 
}

@misc{kim2024openvlaopensourcevisionlanguageactionmodel,
      title={OpenVLA: An Open-Source Vision-Language-Action Model}, 
      author={Moo Jin Kim and Karl Pertsch and Siddharth Karamcheti and Ted Xiao and Ashwin Balakrishna and Suraj Nair and Rafael Rafailov and Ethan Foster and Grace Lam and Pannag Sanketi and Quan Vuong and Thomas Kollar and Benjamin Burchfiel and Russ Tedrake and Dorsa Sadigh and Sergey Levine and Percy Liang and Chelsea Finn},
      year={2024},
      eprint={2406.09246},
      archivePrefix={arXiv},
      primaryClass={cs.RO},
      url={https://arxiv.org/abs/2406.09246}, 
}

@misc{brohan2023rt1roboticstransformerrealworld,
      title={RT-1: Robotics Transformer for Real-World Control at Scale}, 
      author={Anthony Brohan and Noah Brown and Justice Carbajal and Yevgen Chebotar and Joseph Dabis and Chelsea Finn and Keerthana Gopalakrishnan and Karol Hausman and Alex Herzog and Jasmine Hsu and Julian Ibarz and Brian Ichter and Alex Irpan and Tomas Jackson and Sally Jesmonth and Nikhil J Joshi and Ryan Julian and Dmitry Kalashnikov and Yuheng Kuang and Isabel Leal and Kuang-Huei Lee and Sergey Levine and Yao Lu and Utsav Malla and Deeksha Manjunath and Igor Mordatch and Ofir Nachum and Carolina Parada and Jodilyn Peralta and Emily Perez and Karl Pertsch and Jornell Quiambao and Kanishka Rao and Michael Ryoo and Grecia Salazar and Pannag Sanketi and Kevin Sayed and Jaspiar Singh and Sumedh Sontakke and Austin Stone and Clayton Tan and Huong Tran and Vincent Vanhoucke and Steve Vega and Quan Vuong and Fei Xia and Ted Xiao and Peng Xu and Sichun Xu and Tianhe Yu and Brianna Zitkovich},
      year={2023},
      eprint={2212.06817},
      archivePrefix={arXiv},
      primaryClass={cs.RO},
      url={https://arxiv.org/abs/2212.06817}, 
}

@misc{intelligence2025pi06vlalearnsexperience,
      title={$\pi^{*}_{0.6}$: a VLA That Learns From Experience}, 
      author={Physical Intelligence and Ali Amin and Raichelle Aniceto and Ashwin Balakrishna and Kevin Black and Ken Conley and Grace Connors and James Darpinian and Karan Dhabalia and Jared DiCarlo and Danny Driess and Michael Equi and Adnan Esmail and Yunhao Fang and Chelsea Finn and Catherine Glossop and Thomas Godden and Ivan Goryachev and Lachy Groom and Hunter Hancock and Karol Hausman and Gashon Hussein and Brian Ichter and Szymon Jakubczak and Rowan Jen and Tim Jones and Ben Katz and Liyiming Ke and Chandra Kuchi and Marinda Lamb and Devin LeBlanc and Sergey Levine and Adrian Li-Bell and Yao Lu and Vishnu Mano and Mohith Mothukuri and Suraj Nair and Karl Pertsch and Allen Z. Ren and Charvi Sharma and Lucy Xiaoyang Shi and Laura Smith and Jost Tobias Springenberg and Kyle Stachowicz and Will Stoeckle and Alex Swerdlow and James Tanner and Marcel Torne and Quan Vuong and Anna Walling and Haohuan Wang and Blake Williams and Sukwon Yoo and Lili Yu and Ury Zhilinsky and Zhiyuan Zhou},
      year={2025},
      eprint={2511.14759},
      archivePrefix={arXiv},
      primaryClass={cs.LG},
      url={https://arxiv.org/abs/2511.14759}, 
}

@misc{intelligence2026pi07steerablegeneralistrobotic,
      title={${\pi}_{0.7}$: a Steerable Generalist Robotic Foundation Model with Emergent Capabilities}, 
      author={Physical Intelligence and Bo Ai and Ali Amin and Raichelle Aniceto and Ashwin Balakrishna and Greg Balke and Kevin Black and George Bokinsky and Shihao Cao and Thomas Charbonnier and Vedant Choudhary and Foster Collins and Ken Conley and Grace Connors and James Darpinian and Karan Dhabalia and Maitrayee Dhaka and Jared DiCarlo and Danny Driess and Michael Equi and Adnan Esmail and Yunhao Fang and Chelsea Finn and Catherine Glossop and Thomas Godden and Ivan Goryachev and Lachlan Groom and Haroun Habeeb and Hunter Hancock and Karol Hausman and Gashon Hussein and Victor Hwang and Brian Ichter and Connor Jacobsen and Szymon Jakubczak and Rowan Jen and Tim Jones and Gregg Kammerer and Ben Katz and Liyiming Ke and Mairbek Khadikov and Chandra Kuchi and Marinda Lamb and Devin LeBlanc and Brendon LeCount and Sergey Levine and Xinyu Li and Adrian Li-Bell and Vladislav Lialin and Zhonglin Liang and Wallace Lim and Yao Lu and Enyu Luo and Vishnu Mano and Nandan Marwaha and Aikys Mongush and Liam Murphy and Suraj Nair and Tyler Patterson and Karl Pertsch and Allen Z. Ren and Gavin Schelske and Charvi Sharma and Baifeng Shi and Lucy Xiaoyang Shi and Laura Smith and Jost Tobias Springenberg and Kyle Stachowicz and Will Stoeckle and Jiaming Tang and Jimmy Tanner and Shalom Tekeste and Marcel Torne and Kyle Vedder and Quan Vuong and Anna Walling and Haohuan Wang and Jason Wang and XuDong Wang and Chris Whalen and Samuel Whitmore and Blake Williams and Charles Xu and Sukwon Yoo and Lili Yu and Wuming Zhang and Zhuoyang Zhang and Ury Zhilinsky},
      year={2026},
      eprint={2604.15483},
      archivePrefix={arXiv},
      primaryClass={cs.LG},
      url={https://arxiv.org/abs/2604.15483}, 
}

@misc{sun2026autoflyvisionlanguageactionmodeluav,
      title={AutoFly: Vision-Language-Action Model for UAV Autonomous Navigation in the Wild}, 
      author={Xiaolou Sun and Wufei Si and Wenhui Ni and Yuntian Li and Dongming Wu and Fei Xie and Runwei Guan and He-Yang Xu and Henghui Ding and Yuan Wu and Yutao Yue and Yongming Huang and Hui Xiong},
      year={2026},
      eprint={2602.09657},
      archivePrefix={arXiv},
      primaryClass={cs.RO},
      url={https://arxiv.org/abs/2602.09657}, 
}

@misc{zhang2026uavtrackvlaembodiedaerial,
      title={UAV-Track VLA: Embodied Aerial Tracking via Vision-Language-Action Models}, 
      author={Qiyao Zhang and Shuhua Zheng and Jianli Sun and Chengxiang Li and Xianke Wu and Zihan Song and Zhiyong Cui and Yisheng Lv and Yonglin Tian},
      year={2026},
      eprint={2604.02241},
      archivePrefix={arXiv},
      primaryClass={cs.CV},
      url={https://arxiv.org/abs/2604.02241}, 
}

@misc{tucker2026pimakeflyphysicsguided,
      title={$\pi$, But Make It Fly: Physics-Guided Transfer of VLA Models to Aerial Manipulation}, 
      author={Johnathan Tucker and Denis Liu and Aiden Swann and Allen Ren and Javier Yu and Jiankai Sun and Brandon Kim and Lachlain McGranahan and Quan Vuong and Mac Schwager},
      year={2026},
      eprint={2603.25038},
      archivePrefix={arXiv},
      primaryClass={cs.RO},
      url={https://arxiv.org/abs/2603.25038}, 
}

@misc{sun2026airvlavisionlanguageactionsystemsaerial,
      title={AIR-VLA: Vision-Language-Action Systems for Aerial Manipulation}, 
      author={Jianli Sun and Bin Tian and Qiyao Zhang and Chengxiang Li and Zihan Song and Zhiyong Cui and Yisheng Lv and Yonglin Tian},
      year={icml 2026},
      eprint={2601.21602},
      archivePrefix={arXiv},
      primaryClass={cs.RO},
      url={https://arxiv.org/abs/2601.21602}, 
}

@misc{xu2026aerialvlavisionlanguageactionmodeluav,
      title={AerialVLA: A Vision-Language-Action Model for UAV Navigation via Minimalist End-to-End Control}, 
      author={Peng Xu and Zhengnan Deng and Jiayan Deng and Zonghua Gu and Shaohua Wan},
      year={2026},
      eprint={2603.14363},
      archivePrefix={arXiv},
      primaryClass={cs.CV},
      url={https://arxiv.org/abs/2603.14363}, 
}
\clearpage 
\appendix  

\section{Implementation and Training Details}
\label{app:implementation}

\textbf{Architecture Specifications:} 
Our proposed AIR-VLA+ builds upon the Physical Intelligence $\pi_{0.5}$ foundation model as the primary vision-language backbone. To construct the \textit{input feature enhancement module}, we apply mean pooling over the sequence output of the prefix tokens (images and text) to derive a highly condensed global context representation, which is then temporally broadcasted across the action sequence horizon. The \textit{implicit visual grasp projector} is instantiated as an MLP consisting of two linear layers with a ReLU activation, squeezing the latent dimension from the backbone width (256 dimensions) down to 128 dimensions to enforce physical state discriminability. 

For the \textit{unidirectional information transmission cascade channel}, the fused input for the UAV movement decoder concatenates the visual features, the broadcasted global context, the visual grasp features, and the intended action from the arm manipulation decoder. Crucially, a \textit{stop-gradient operation} (\texttt{.detach()}) is explicitly applied to the manipulator's intended action tensor before concatenation, ensuring strict isolation of backpropagation dynamics. The \textit{feature-level MoE} routing network utilizes a 2-layer MLP (hidden dimension 128) outputting through a Softmax function to distribute weights across $N=3$ linear movement experts. Finally, \textit{physical hard isolation} is strictly enforced on the action dimension indices, mapping the first four dimensions strictly to the UAV movement decoder and the remaining dimensions to the arm manipulation decoder.

\textbf{Two-Stage Training Paradigm:} 
To prevent catastrophic forgetting of the foundational physical priors and stabilize the MoE routing, we devised a two-stage training pipeline.
\begin{itemize}
    \item \textbf{Stage 1 (Visual Cascade Adaptation):} The network is initialized from the base $\pi_{0.5}$ weights. The entire framework, including the newly added visual extractor and asymmetric dual decoders, is trained for 30,000 steps. We employ the AdamW optimizer with a peak learning rate of 5e-5, warming up over 10,000 steps, and applying a cosine decay schedule.
    \item \textbf{Stage 2 (Feature MoE Fine-Tuning):} Initialized from the Stage 1 checkpoint, we freeze the vast majority of the foundational multimodal backbone to protect the representation manifold. Gradients are selectively unblocked only for the MoE routing network, the $N=3$ movement experts, the visual grasp projector, and the manipulation projection decoder. This stage trains for an additional 10,000 steps with a reduced peak learning rate of 2e-5 (warmup of 2,000 steps, decaying to 1e-5).
\end{itemize}

\textbf{Hardware, Inference, and Evaluation Setup:} 
All training stages were conducted using Distributed Data Parallel (DDP) across 2 NVIDIA A100 GPUs, maintaining a global effective batch size of 32. Gradients were clipped at a maximum norm of 1.0 to ensure stability. During the inference phase, the continuous trajectory generation is solved using an Euler ODE solver, configured with 10 Neural Function Evaluations (NFE) uniformly spaced across the integration time horizon $t \in [0, 1]$. Finally, to ensure statistical reliability during evaluation, the reported performance metrics for each individual task are calculated as the average over 20 independent rollout trials.
\end{document}